\documentclass[11pt,a4paper]{article}
\usepackage[hyperref]{acl2017}
\usepackage{times}
\usepackage{latexsym}

\usepackage{url}

\usepackage{amsmath}
\usepackage{graphicx}
\usepackage{multirow}
\usepackage[export]{adjustbox}
\usepackage{booktabs}
\usepackage[ruled,norelsize]{algorithm2e}

\aclfinalcopy 

\title{Extract with Order for Coherent Multi-Document Summarization}

\author{Mir Tafseer Nayeem \\
  University of Lethbridge  \\
  Lethbridge, AB, Canada  \\
  {\tt mir.nayeem@uleth.ca} \\\And
  Yllias Chali \\
  University of Lethbridge  \\
  Lethbridge, AB, Canada \\
  {\tt chali@cs.uleth.ca} \\}

\date{}

\begin{document}
\maketitle
\begin{abstract}
 In this work, we aim at developing an extractive summarizer in the multi-document setting. We implement a rank based sentence selection using continuous vector representations along with key-phrases. Furthermore, we propose a model to tackle summary coherence for increasing readability.  We conduct experiments on the Document Understanding Conference (DUC) 2004 datasets using ROUGE toolkit. Our experiments demonstrate that the methods bring significant improvements over the state of the art methods in terms of informativity and coherence.
\end{abstract}

\section{Introduction}

The task of automatic document summarization aims at finding the most relevant informations in a text and presenting them in a condensed form. A good summary should retain the most important contents of the original document  or a cluster of documents, while being coherent, non-redundant and grammatically readable. There are two types of summarizations: abstractive summarization and extractive summarization. Abstractive methods,  which are still a growing field are highly complex as they need extensive natural language generation to rewrite the sentences \cite{DBLP:conf/cikm/NayeemC17, DBLP:conf/ijcnlp/ChaliTN17}. Therefore, research community is focusing more on extractive summaries, which selects salient (important) sentences from the source document without any modification to create a summary \cite{nayeem-etal-2018-abstractive}. Summarization is classified as single-document or multi-document based upon the number of source document.  The information overlap between the documents from the same topic makes the multi-document summarization more challenging than the task of summarizing single documents.

One crucial step in generating a coherent summary is to order the sentences in a logical manner to increase the readability. A wrong order of sentences convey entirely different idea to the reader of the summary and also make it difficult to understand. In a single document, summary information can be presented by preserving the sentence position in the original document. In multi-document summarization, the sentence position in the original document does not provide clue to the sentence arrangement. Hence it is a very challenging task to perform the arrangement of sentences in the summary.

\section{Related Work}
During a decade, several extractive approaches have been developed for automatic summary generation that implement a number of machine learning, graph-based and optimization techniques. LexRank \cite{Erkan:2004:LGL:1622487.1622501} and TextRank \cite{mihalcea-tarau:2004:EMNLP} are graph-based methods of
computing sentence importance for text summarization. The RegSum system \cite{hong-nenkova:2014:EACL} employs a supervised model for predicting word importance. Treating multi-document summarization as a submodular maximization problem has proven successful by \cite{Lin:2011:CSF:2002472.2002537}. Unfortunately, none of the above systems care about the coherence of the final extracted summary.

In very recent works using neural network, \cite{cheng-lapata:2016:P16-1} proposed an attentional encoder-decoder and \cite{AAAI1714636} used a simple recurrent network based sequence classifier to solve the problem of extractive summarization. However, they are limited to single document settings, where sentences are implicitly ordered according to the sentence position. \cite{Parveen:2015:IIN:2832415.2832430, parveen-ramsl-strube:2015:EMNLP} proposed  graph-based techniques to tackle coherence, which is also limited to single document summarization. Moreover, a recent work \cite{wang-EtAl:2016:COLING1} actually proposed a multi-document summarization system that combines both coherence and informativeness but this system is limited to syntactic linkages between entities.

In this paper, we implement a rank based sentence selection using  continuous  vector representations  along  with  key-phrases. We  also model the coherence using semantic relations between entities and sentences to increase the readability. 

\section{Sentence Extraction}

We here successively describe each of the steps involved in the sentence extraction process such as sentence ranking, sentence clustering, and sentence selection. 

\subsection{Preprocessing}

Our system first takes a set of related texts as input and preprocesses them which includes tokenization, Part-Of-Speech (POS) tagging, removal of stopwords and Lemmatization. We use \textbf{NLTK} toolkit\footnote{http://www.nltk.org/} to preprocess each sentence to obtain a more accurate representation of the information.

\subsection{Sentence Similarity}

 We take the pre-trained word embeddings\footnote{https://code.google.com/archive/p/word2vec/} \cite{Mikolov:2013:DRW:2999792.2999959}  of all the non stopwords in a sentence and take the weighted vector sum according to the term-frequency ($TF$) of a word($w$) in a sentence($S$). Where, $E$ is the word embedding model and $idx(w)$ is the index of the word $w$. More formally, for a given sentence $ S$ in the document $D$, the weighted sum becomes,
   
   $$ S = \sum_{ w \in {S}} TF(w,S)\cdot E[idx(w)] $$
   
Then we calculate cosine similarity between the sentence vectors obtained from the above equation to find the relative distance between $S_i$ and $S_j$. We also calculate $ NESim(S_i , S_j)$ by finding the Named Entities present in $S_i$ and $S_j$ using NLTK Toolkit, then calculating their overlap.
   
$$ CosSim(S_i , S_j) = \frac {S_i \cdot S_j} {||S_i|| \ ||S_j||} $$
   
$$ NESim(S_i , S_j) = \frac {| NE(S_i) \cap NE(S_j) |} { min( | NE(S_i)| ,|NE(S_j)|)} $$
 \begin{multline}
 Sim(S_i , S_j) = \lambda \cdot NESim(S_i , S_j) \ +  \\ (1 - \lambda) \cdot CosSim(S_i , S_j)
 \end{multline}

The overall similarity calculation involves both $ CosSim(S_i , S_j)$ and $ NESim(S_i , S_j)$ where, $ 0 \leq \lambda \leq 1$ decides the relative contributions of them to the overall similarity computation. This standalone similarity function will be used in this work with different $\lambda$ values to accomplish different tasks.

\subsection{Sentence Ranking}

In this section, we rank the sentences by applying TextRank algorithm \cite{mihalcea-tarau:2004:EMNLP} which involves constructing an undirected graph where sentences are vertices, and weighted edges are formed connecting sentences by a similarity metric. TextRank determines the similarity based on the lexical overlap  between two sentences. However, this algorithm has a serious drawback: If two sentences are talking about the same topic without using any overlapped words, there will be no edge between them. Instead, we use the continuous skip-gram model introduced by \cite{ Mikolov:2013:DRW:2999792.2999959} to measure the semantic similarity along with the entity overlap. We use the similarity function described in Equation (1) by setting $ \lambda = 0.3 $.

After we have our graph, we can run the main algorithm on it. This involves initializing a score of 1 for each vertex, and repeatedly applying the TextRank update rule until convergence. The update rule is:
\begin{multline*}
    Rank(S_i) = (1 - d) + d \ * \\ \sum_{S_j \in N(S_i)} \frac{Sim(S_i,S_j)}{\sum_{S_k \in N(S_j)} Sim(S_j,S_k)} Rank(S_j)
\end{multline*}

Where, $Rank(S_i)$ indicates the importance score assigned to sentence $S_i$. $N(S_i)$ is the set of neighboring sentences of $S_i$, and $0 \leq d \leq 1$ is a dampening factor, which the literature suggests its setting to 0.85. After reaching convergence, we extract the sentences along with TextRank scores.

\subsection{Sentence Clustering}

The sentence clustering step allows us to group similar sentences. We use a hierarchical agglomerative clustering \cite{Murtagh:2014:WHA:2689486.2689593} with a complete linkage criteria. This method proceeds incrementally, starting with each sentence considered as a cluster, and merging the pair of similar clusters after each step using bottom up approach.  The complete linkage criteria determines the metric used for the merge strategy. In computing the clusters, we use the similarity function described in Equation (1) with $ \lambda = 0.4 $. We set a similarity threshold ($ \tau = 0.5 $)  to stop the clustering process. If we cannot find any cluster pair with a similarity above the threshold, the process stops, and the clusters are released. The clusters may be small, but are highly coherent as each sentence they contain must be similar to every other sentence in the same cluster. 

This sentence clustering step is very important due to two main reasons, (1) Selecting at most one sentence from each cluster of related sentences will decrease redundancy from the summary side (2) Selecting sentences from the diverse set of clusters will increase the information coverage from the document side as well.

\subsection{Sentence Selection}

In this work, we use the concept-based ILP framework introduced in \cite{Gillick:2009:SGM:1611638.1611640} with some suitable changes to select the best subset of sentences. This approach aims to extract sentences that cover as many important concepts as possible, while ensuring the summary length is within a given budgeted constraint. Unlike \cite{Gillick:2009:SGM:1611638.1611640} which uses bigrams as concepts, we use keyphrases as concepts. Keyphrases are the words or phrases that represent the main topics of a document. Sentences containing the
most relevant keyphrases are important for the summary generation. We extracted the keyphrases from the document cluster using RAKE\footnote{https://github.com/aneesha/RAKE} \cite{rose2010automatic}. We assign a weight to each keyphrase using the score returned by RAKE.

Let $w_i$ be the weight of keyphrase $i$ and $k_i$ a binary variable that indicates the presence of keyphrase $i$ in the extracted sentences. Let $l_j$ be the number of words in sentence j, $s_j$ a binary variable that indicates the presence of sentence $j$ in the extracted sentence set and $L$ the length limit for the set. Let $Occ_{ij}$ indicate the occurrence of keyphrase $i$ in sentence $j$, the ILP formulation is,

   \begin{equation}
   Maximize: ( \sum_{i} w_i k_i + \sum_{j} Rank(S_j) \cdot s_j )
   \end{equation}
   
   \begin{equation}
   Subject\ to: \sum_{j} l_j s_j \leq L
   \end{equation}
   
   \begin{equation}
    s_j Occ_{ij} \leq k_i, \quad \forall{i,j} 
   \end{equation}
   
   \begin{equation}
   \sum_{j} s_j Occ_{ij} \geq k_i, \quad \forall{i} 
   \end{equation}
   
   \begin{equation}
   \sum_{j \in {g_c}} s_j \leq 1, \quad \forall{g_c}
   \end{equation}
   
   \begin{equation}
     k_i \in \{0,1\}  \quad \forall{i} 
   \end{equation}
   
   \begin{equation}
    s_j \in \{0,1\}  \quad \forall{j} 
   \end{equation}

We try to maximize the weight of the keyphrases (2) in the extracted sentences, while avoiding repetition of those keyphrases (4, 5) and staying under the maximum number of words allowed for the sentence extraction (3).

In addition to \cite{Gillick:2009:SGM:1611638.1611640}, we put some extra features like maximizing the sentence rank scores returned from the sentence ranking section. In order to ensure only one sentence per cluster in the extracted sentences we add an extra constraint (6). In this process, we extract the optimal combination of sentences that maximize informativity while minimizing redundancy (Figure 1 illustrates our sentence extraction process in brief).

\begin{figure}[ht]
\includegraphics[width=.43\textwidth,center]{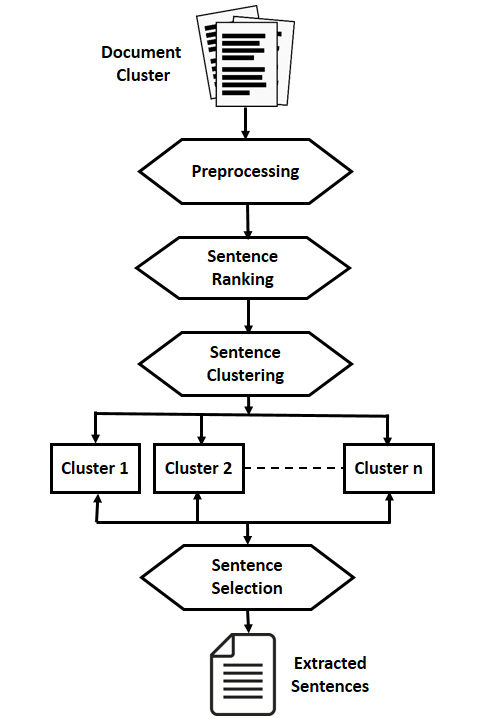}
\caption{Sentence Extraction Process}
\end{figure}

\begin{table*}
\centering
\small
\label{my-label}
\begin{tabular}{|c|c|c|c|l|c|}
\hline
\textbf{System} & \textbf{Models} & \textbf{R-1} & \textbf{R-2} & \textbf{R-SU4} & \textbf{Coherence} \\ \hline
\multirow{2}{*}{\textbf{Baseline}} & LexRank & 35.95 & 7.47 & 12.48 & 0.39 \\ \cline{2-6} 
 & GreedyKL & 37.98 & 8.53 & 13.25 & 0.46 \\ \hline
\multirow{2}{*}{\textbf{State-of-the-art}} & Submodular & 39.18 & 9.35 & \textbf{14.22} & 0.51 \\ \cline{2-6} 
 & ICSISumm & 38.41 & 9.78 & 13.31 & 0.44 \\ \hline
\textbf{Proposed System} & ILPRankSumm & \textbf{39.45} & \textbf{10.12} & 14.09 & \textbf{0.68} \\ \hline
\end{tabular}
\caption{ Results on DUC 2004 (Task-2) for the baseline, state-of-the-art and
our system.}
\end{table*}

\section{Sentence Ordering}

Classic reordering approaches include inferring order from weighted sentence graph \cite{Barzilay:2002:ISS:1622810.1622812}, or perform a chronological ordering algorithm \cite{Cohen:1999:LOT:1622859.1622867} that sorts sentences based on timestamp and position.

We here propose a simple greedy approach to sentence ordering in multi-document settings.  Our assumption is that a good sentence order implies the similarity between all adjacent sentences since word repetition (more specifically, named entity repetition) is one of the formal sign of text coherence \cite{Barzilay:2002:ISS:1622810.1622812}. We define coherence of document $D$ which consists of sentences from $S_1$ to $S_n$ in the following equation. For calculating $Sim(S_i \ ,S_{i+1})$, we use the similarity function described in equation (1) with $\lambda = 0.5$, giving the named entities a little more preference.

$$ Coherence (D) = \frac { \sum_{i=1}^{n-1} Sim(S_i \ ,S_{i+1})} {n-1} $$

We propose a greedy algorithm for placing a sentence in a document based on the coherence score we discussed above\footnote{Note that, we didn't take any position information of the original sentences to be extracted from the document.}. At the beginning, we randomly select a sentence from the extracted sentences without any position information and place the sentence in the ordered set $D$. We then incrementally add each extracted sentences to the document set $D$ using Algorithm (1) to get the final order of summary sentences.

\begin{algorithm}[ht]
\small
\SetAlgoLined
\SetKwFunction{SentencePositioning}{SentencePositioning}
\SetKwProg{myproc}{Procedure}{}{}
\myproc{\SentencePositioning{$D,S_n$}}{
\KwData{Input document $D$ which is assumed sorted. New sentence $S_n$ which we will place in the document $D$.}
\KwResult{Return new document $D_n$ after placing the sentence $S_n$. }
 $ t \leftarrow 1 $\;
 $ Coh_{max} \leftarrow 0 $ \;
 $ D_{tmp} \leftarrow D $ \;
 $ l \leftarrow DocLength(D) $ \;
 \While{ $t \leq l + 1 $}{
  \textit{$\Rightarrow$Place the $S_n$ in $t^{th}$ position of $D_{tmp}$ }\;
    $ Coh_{tmp} \leftarrow Coherence(D_{tmp})$\;
  \If{$Coh_{tmp} >  Coh_{max} $}{
   $D_n \leftarrow D_{tmp} $\;
   $Coh_{max} \leftarrow Coh_{tmp} $\;
   \textit{$\Rightarrow$ Remove $S_n$ from the $t^{th}$ position of the document $D_{tmp}$ }\;
   }
 $ t \leftarrow t + 1 $\;
 }
 \KwRet $D_n$\;}
 \caption{Place a sentence to a document }
\end{algorithm}

\section{Evaluation}

We evaluate our system \textbf{ILPRankSumm} (\textbf{ILP} based sentence selection with Text\textbf{Rank} for Extractive \textbf{Summ}arization) using \textbf{ROUGE}\footnote{ROUGE-1.5.5 with options: -n 2 -m -u -c 95 -x -r 1000 -f A -p 0.5 -t 0} \cite{lin:2004:ACLsummarization} on DUC 2004 (Task-2, Length limit($L$) = 100 words). However, ROUGE scores are biased towards lexical overlap at surface level and insensitive to summary coherence. Moreover, sophisticated coherence evaluation metrics are seldom adopted for summarization thus many of the previous systems used human evaluation for measuring readability. For this reason, we evaluate our summary coherence using \cite{Lapata:2005:AET:1642293.1642467} \cite{Barzilay:2008:MLC:1350986.1350987} which defines coherence probabilities for an ordered set of sentences. 

\subsection{Baseline Systems}
We compare our system with baseline (LexRank, GreedyKL) and state of the art systems (Submodular, ICSISumm). \textbf{LexRank}\cite{Erkan:2004:LGL:1622487.1622501} represents input texts as graph where nodes are the sentences and the edges are formed between two sentences if the cosine similarity is above a certain threshold. Sentence importance is calculated by running the PageRank algorithm on the graph. \textbf{GreedyKL} \cite{Haghighi:2009:ECM:1620754.1620807} iteratively selects the next sentence for the summary that will minimize the KL divergence between the estimated word distributions.
\cite{Lin:2011:CSF:2002472.2002537} treat the document summarization problem as maximizing a \textbf{Submodular} function under a budget constraint. They achieved a near-optimal information coverage and non-redundancy using a modified greedy algorithm. On the other hand, \textbf{ICSISumm} \cite{Gillick:2009:SGM:1611638.1611640} employs a global linear optimization framework, finding the globally optimal summary rather than choosing sentences according to their importance in a greedy fashion.

The summaries generated by the baselines and the state-of-the-art extractive summarizers on the DUC 2004 dataset were collected from \cite{HONG14.1093.L14-1070}.

\subsection{Results}
Our results include R-1, R-2, and R-SU4, which counts matches in unigrams, bigrams, and skip-bigrams respectively. The skip-bigrams allow four words
in between. According to Table 1, R-1, R-2 scores obtained by our system outperform all the baselines and state of the art systems on DUC 2004 datasets. One of the main reasons of getting the improved R-1 and R-2 score is the use of keyphrases. Moreover, there is no significant difference between our proposed system and submodular in case of R-SU4. We also get better coherence probability because of our sentence ordering technique. The system's output for a randomly selected document set (e.g. d30015t) from DUC 2004 is shown in Table 2.

\subsection{Limitations}
One of the essential properties of the text summarization systems is the ability to generate a summary with a fixed length (DUC 2004, Task-2: Length limit = 100 words). According to \cite{HONG14.1093.L14-1070} all the summarizer from the previous research either truncated the summary to $100^{th}$ word, or removed the last sentence from the summary set. In this paper, we follow the second one to produce grammatical summary. However, the first one produces a certain ungrammatical sentence, later one can lose a lot of information in the worst case, if the sentences are long. We more focus on the grammaticality of the final summary.

\begin{table}[]
\centering
\small
\label{my-label}
\begin{tabular}{|l|}
\hline
\multicolumn{1}{|c|}{\textbf{Summary Generated (After Sentence Extraction)}}                                                                                                                                                                                                                                                                                                                                                                                                                                                                                                                                                                                                                                                                    \\ \hline
\begin{tabular}[c]{@{}l@{}}But U.S. special envoy Richard Holbrooke said the \\ situation in the southern Serbian province was as bad \\ now as two weeks ago. A Western diplomat said up to \\ 120 Yugoslav army armored vehicles, including tanks, \\ have been pulled out. On Sunday,Milosevic met with \\ Russian Foreign Minister Igor Ivanov and Defense \\ Minister Igor Sergeyev, Serbian President Milan \\ Milutinovic and Yugoslavia's top defense officials. \\ To avoid such an attack, Yugoslavia must end the \\ hostilities, withdraw army and security forces, take \\ urgent measures to overcome the humanitarian crisis, \\ ensure that refugees can return home and take part \\ in peace talks, he said.\end{tabular}     \\ \hline
\multicolumn{1}{|c|}{\textbf{Summary Generated (After Sentence Ordering)}}                                                                                                                                                                                                                                                                                                                                                                                                                                                                                                                                                                                                                                                                      \\ \hline
\begin{tabular}[c]{@{}l@{}}On Sunday, Milosevic met with Russian Foreign \\ Minister Igor Ivanov and Defense Minister Igor \\ Sergeyev, Serbian President Milan Milutinovic and \\ Yugoslavia's top defense officials. But U.S. special \\ envoy Richard Holbrooke said the situation in the \\ southern Serbian province was as bad now as two \\ weeks ago. A Western diplomat said up to 120 \\ Yugoslav army armored vehicles, including tanks, \\ have been pulled out. To avoid such an attack, \\ Yugoslavia must end the hostilities, withdraw army \\ and security forces, take urgent measures to \\ overcome the humanitarian crisis, ensure that \\ refugees can return home and take part in peace talks, \\ he said.\end{tabular} \\ \hline
\end{tabular}
\caption{System's output (100 words) for the document set \textbf{d30015t} from DUC 2004.}
\end{table}

\section{Conclusion and Future Work}

In this work, we implemented an ILP based sentence selection along with TextRank scores and key phrases for extractive multi-document summarization. We further model the coherence to increase the  readability of the generated summary. Evaluation results strongly indicate the benefits of using continuous word vector representations in all the steps involved in the overall system. In future, we will focus on jointly extracting the sentences to maximize informativity and readability while minimizing redundancy using the same ILP model. Moreover, we will also try to propose a solution for the length limit problem.

\section*{Acknowledgments}

We would like to thank the anonymous reviewers for their useful comments. The research reported in this paper was conducted at the
University of Lethbridge and supported by the Natural Sciences and Engineering Research Council (NSERC) of Canada – discovery grant and the University of Lethbridge.

\bibliography{acl2017}

\begin{thebibliography}{}
\expandafter\ifx\csname natexlab\endcsname\relax\def\natexlab#1{#1}\fi

\bibitem[{Barzilay et~al.(2002)Barzilay, Elhadad, and
  McKeown}]{Barzilay:2002:ISS:1622810.1622812}
Regina Barzilay, Noemie Elhadad, and Kathleen~R. McKeown. 2002.
\newblock Inferring strategies for sentence ordering in multidocument news
  summarization.
\newblock {\em J. Artif. Int. Res.\/} 17(1):35--55.

\bibitem[{Barzilay and Lapata(2008)}]{Barzilay:2008:MLC:1350986.1350987}
Regina Barzilay and Mirella Lapata. 2008.
\newblock Modeling local coherence: An entity-based approach.
\newblock {\em Comput. Linguist.\/} 34(1):1--34.

\bibitem[{Chali et~al.(2017)Chali, Tanvee, and
  Nayeem}]{DBLP:conf/ijcnlp/ChaliTN17}
Yllias Chali, Moin Tanvee, and Mir~Tafseer Nayeem. 2017.
\newblock Towards abstractive multi-document summarization using submodular
  function-based framework, sentence compression and merging.
\newblock In {\em Proceedings of the Eighth International Joint Conference on
  Natural Language Processing, {IJCNLP} 2017, Taipei, Taiwan, November 27 -
  December 1, 2017, Volume 2: Short Papers\/}. pages 418--424.

\bibitem[{Cheng and Lapata(2016)}]{cheng-lapata:2016:P16-1}
Jianpeng Cheng and Mirella Lapata. 2016.
\newblock Neural summarization by extracting sentences and words.
\newblock In {\em Proceedings of the 54th Annual Meeting of the Association for
  Computational Linguistics (Volume 1: Long Papers)\/}. Association for
  Computational Linguistics, Berlin, Germany, pages 484--494.

\bibitem[{Cohen et~al.(1999)Cohen, Schapire, and
  Singer}]{Cohen:1999:LOT:1622859.1622867}
William~W. Cohen, Robert~E. Schapire, and Yoram Singer. 1999.
\newblock Learning to order things.
\newblock {\em J. Artif. Int. Res.\/} 10(1):243--270.

\bibitem[{Erkan and Radev(2004)}]{Erkan:2004:LGL:1622487.1622501}
G\"{u}nes Erkan and Dragomir~R. Radev. 2004.
\newblock Lexrank: Graph-based lexical centrality as salience in text
  summarization.
\newblock {\em J. Artif. Int. Res.\/} 22(1):457--479.

\bibitem[{Gillick and Favre(2009)}]{Gillick:2009:SGM:1611638.1611640}
Dan Gillick and Benoit Favre. 2009.
\newblock A scalable global model for summarization.
\newblock In {\em Proceedings of the Workshop on Integer Linear Programming for
  Natural Langauge Processing\/}. Association for Computational Linguistics,
  Stroudsburg, PA, USA, ILP '09, pages 10--18.

\bibitem[{Haghighi and Vanderwende(2009)}]{Haghighi:2009:ECM:1620754.1620807}
Aria Haghighi and Lucy Vanderwende. 2009.
\newblock Exploring content models for multi-document summarization.
\newblock In {\em Proceedings of Human Language Technologies: The 2009 Annual
  Conference of the North American Chapter of the Association for Computational
  Linguistics\/}. Association for Computational Linguistics, Stroudsburg, PA,
  USA, NAACL '09, pages 362--370.

\bibitem[{Hong et~al.(2014)Hong, Conroy, Favre, Kulesza, Lin, and
  Nenkova}]{HONG14.1093.L14-1070}
Kai Hong, John Conroy, Benoit Favre, Alex Kulesza, Hui Lin, and Ani Nenkova.
  2014.
\newblock A repository of state of the art and competitive baseline summaries
  for generic news summarization.
\newblock In {\em Proceedings of the Ninth International Conference on Language
  Resources and Evaluation (LREC'14)\/}. European Language Resources
  Association (ELRA), Reykjavik, Iceland, pages 1608--1616.
\newblock ACL Anthology Identifier: L14-1070.

\bibitem[{Hong and Nenkova(2014)}]{hong-nenkova:2014:EACL}
Kai Hong and Ani Nenkova. 2014.
\newblock Improving the estimation of word importance for news multi-document
  summarization.
\newblock In {\em Proceedings of the 14th Conference of the European Chapter of
  the Association for Computational Linguistics\/}. Association for
  Computational Linguistics, Gothenburg, Sweden, pages 712--721.

\bibitem[{Lapata and Barzilay(2005)}]{Lapata:2005:AET:1642293.1642467}
Mirella Lapata and Regina Barzilay. 2005.
\newblock Automatic evaluation of text coherence: Models and representations.
\newblock In {\em Proceedings of the 19th International Joint Conference on
  Artificial Intelligence\/}. Morgan Kaufmann Publishers Inc., San Francisco,
  CA, USA, IJCAI'05, pages 1085--1090.

\bibitem[{Lin(2004)}]{lin:2004:ACLsummarization}
Chin-Yew Lin. 2004.
\newblock Rouge: A package for automatic evaluation of summaries.
\newblock In Stan~Szpakowicz Marie-Francine~Moens, editor, {\em Text
  Summarization Branches Out: Proceedings of the ACL-04 Workshop\/}.
  Association for Computational Linguistics, Barcelona, Spain, pages 74--81.

\bibitem[{Lin and Bilmes(2011)}]{Lin:2011:CSF:2002472.2002537}
Hui Lin and Jeff Bilmes. 2011.
\newblock A class of submodular functions for document summarization.
\newblock In {\em Proceedings of the 49th Annual Meeting of the Association for
  Computational Linguistics: Human Language Technologies - Volume 1\/}.
  Association for Computational Linguistics, Stroudsburg, PA, USA, HLT '11,
  pages 510--520.

\bibitem[{Mihalcea and Tarau(2004)}]{mihalcea-tarau:2004:EMNLP}
Rada Mihalcea and Paul Tarau. 2004.
\newblock Textrank: Bringing order into texts.
\newblock In Dekang Lin and Dekai Wu, editors, {\em Proceedings of EMNLP
  2004\/}. Association for Computational Linguistics, Barcelona, Spain, pages
  404--411.

\bibitem[{Mikolov et~al.(2013)Mikolov, Sutskever, Chen, Corrado, and
  Dean}]{Mikolov:2013:DRW:2999792.2999959}
Tomas Mikolov, Ilya Sutskever, Kai Chen, Greg Corrado, and Jeffrey Dean. 2013.
\newblock Distributed representations of words and phrases and their
  compositionality.
\newblock In {\em Proceedings of the 26th International Conference on Neural
  Information Processing Systems\/}. Curran Associates Inc., USA, NIPS'13,
  pages 3111--3119.

\bibitem[{Murtagh and Legendre(2014)}]{Murtagh:2014:WHA:2689486.2689593}
Fionn Murtagh and Pierre Legendre. 2014.
\newblock Ward's hierarchical agglomerative clustering method: Which algorithms
  implement ward's criterion?
\newblock {\em J. Classif.\/} 31(3):274--295.

\bibitem[{Nallapati et~al.(2017)Nallapati, Zhai, and Zhou}]{AAAI1714636}
Ramesh Nallapati, Feifei Zhai, and Bowen Zhou. 2017.
\newblock Summarunner: {A} recurrent neural network based sequence model for
  extractive summarization of documents.
\newblock In {\em Proceedings of the Thirty-First {AAAI} Conference on
  Artificial Intelligence, February 4-9, 2017, San Francisco, California,
  {USA.}\/}. pages 3075--3081.

\bibitem[{Nayeem and Chali(2017)}]{DBLP:conf/cikm/NayeemC17}
Mir~Tafseer Nayeem and Yllias Chali. 2017.
\newblock Paraphrastic fusion for abstractive multi-sentence compression
  generation.
\newblock In {\em Proceedings of the 2017 {ACM} on Conference on Information
  and Knowledge Management, {CIKM} 2017, Singapore, November 06 - 10, 2017\/}.
  pages 2223--2226.

\bibitem[{Nayeem et~al.(2018)Nayeem, Fuad, and
  Chali}]{nayeem-etal-2018-abstractive}
Mir~Tafseer Nayeem, Tanvir~Ahmed Fuad, and Yllias Chali. 2018.
\newblock Abstractive unsupervised multi-document summarization using
  paraphrastic sentence fusion.
\newblock In {\em Proceedings of the 27th International Conference on
  Computational Linguistics\/}. Association for Computational Linguistics,
  Santa Fe, New Mexico, USA, pages 1191--1204.

\bibitem[{Parveen et~al.(2015)Parveen, Ramsl, and
  Strube}]{parveen-ramsl-strube:2015:EMNLP}
Daraksha Parveen, Hans-Martin Ramsl, and Michael Strube. 2015.
\newblock Topical coherence for graph-based extractive summarization.
\newblock In {\em Proceedings of the 2015 Conference on Empirical Methods in
  Natural Language Processing\/}. Association for Computational Linguistics,
  Lisbon, Portugal, pages 1949--1954.

\bibitem[{Parveen and Strube(2015)}]{Parveen:2015:IIN:2832415.2832430}
Daraksha Parveen and Michael Strube. 2015.
\newblock Integrating importance, non-redundancy and coherence in graph-based
  extractive summarization.
\newblock In {\em Proceedings of the 24th International Conference on
  Artificial Intelligence\/}. AAAI Press, IJCAI'15, pages 1298--1304.

\bibitem[{Rose et~al.(2010)Rose, Engel, Cramer, and Cowley}]{rose2010automatic}
Stuart Rose, Dave Engel, Nick Cramer, and Wendy Cowley. 2010.
\newblock Automatic keyword extraction from individual documents.
\newblock {\em Text Mining\/} pages 1--20.

\bibitem[{Wang et~al.(2016)Wang, Nishino, Hirao, Sudoh, and
  Nagata}]{wang-EtAl:2016:COLING1}
Xun Wang, Masaaki Nishino, Tsutomu Hirao, Katsuhito Sudoh, and Masaaki Nagata.
  2016.
\newblock Exploring text links for coherent multi-document summarization.
\newblock In {\em Proceedings of COLING 2016, the 26th International Conference
  on Computational Linguistics: Technical Papers\/}. The COLING 2016 Organizing
  Committee, Osaka, Japan, pages 213--223.

\end{thebibliography}
\bibliographystyle{acl_natbib}

\appendix

\end{document}